\documentclass[a4paper,conference]{IEEEtran}
\usepackage{cite}
\usepackage{amsmath,amssymb,amsfonts}
\usepackage{algorithmic}
\usepackage{graphicx}

\ifCLASSOPTIONcompsoc
 \usepackage[caption=false,font=normalsize,labelfont=sf,textfont=sf]{subfig}
\else
 \usepackage[caption=false,font=footnotesize]{subfig}
\fi
    
\usepackage{booktabs}
\usepackage{adjustbox}
\usepackage{tikz}  

\usepackage{amsthm}
\theoremstyle{definition}
\newtheorem{property}{Property}

\newcommand{\erics}{\emph{ERICS}}

\newcommand\copyrighttext{%
  \footnotesize \textcopyright 2020 IEEE. Personal use of this material is permitted. Permission from IEEE must be obtained for all other uses, in any current or future media, including reprinting/republishing this material for advertising or promotional purposes, creating new collective works, for resale or redistribution to servers or lists, or reuse of any copyrighted component of this work in other works.}
\newcommand\copyrightnotice{%
\begin{tikzpicture}[remember picture,overlay]
\node[anchor=south,yshift=10pt] at (current page.south) {\fbox{\parbox{\dimexpr\textwidth-\fboxsep-\fboxrule\relax}{\copyrighttext}}};
\end{tikzpicture}%
}

\begin{document}

\title{Learning Parameter Distributions to Detect Concept Drift in Data Streams}

\author{\IEEEauthorblockN{Johannes Haug}
\IEEEauthorblockA{University of Tuebingen \\
Tuebingen, Germany \\
johannes-christian.haug@uni-tuebingen.de}
\and
\IEEEauthorblockN{Gjergji Kasneci}
\IEEEauthorblockA{University of Tuebingen \\
Tuebingen, Germany \\
gjergji.kasneci@uni-tuebingen.de}
}

\maketitle
\copyrightnotice  

\begin{abstract}
Data distributions in streaming environments are usually not stationary. In order to maintain a high predictive quality at all times, online learning models need to adapt to distributional changes, which are known as concept drift. The timely and robust identification of concept drift can be difficult, as we never have access to the true distribution of streaming data. In this work, we propose a novel framework for the detection of real concept drift, called \erics. By treating the parameters of a predictive model as random variables, we show that concept drift corresponds to a change in the distribution of optimal parameters. To this end, we adopt common measures from information theory. The proposed framework is completely model-agnostic. By choosing an appropriate base model, \erics\ is also capable to detect concept drift at the input level, which is a significant advantage over existing approaches. An evaluation on several synthetic and real-world data sets suggests that the proposed framework identifies concept drift more effectively and precisely than various existing works.
\end{abstract}


\section{Introduction}\label{sec:introduction}
Data streams are a potentially unbounded sequence of observations. As such, data streams are subject to a number of external factors, e.g. seasonal or catastrophic events. Hence, the distributions of a data stream are usually not stationary, but change over time, which is known as concept drift.

Concept drift can seriously affect the quality of predictions, if it goes unnoticed. Concept drift detection models help identify and handle distributional changes, allowing us to maintain a high predictive performance over time. 
Ideally, concept drift detection models are sensitive enough to detect drift with only a short delay. However, concept drift detection should also be robust against small perturbations of the input in order to avoid false positives and thus be reliable.

Let $X$ and $Y$ be random variables that correspond to the streaming observations and the associated labels. According to \cite{webb2016characterizing}, concept drift resembles a difference in the joint probability $P(Y,X)$ at different time steps $t, u \in \{1,..,\mathcal{T}\}$, i.e.
\begin{align*}
    P_t(Y,X) &\neq P_u(Y,X)\\
    \Leftrightarrow P_t(Y|X)P_t(X) &\neq P_u(Y|X)P_u(X).
\end{align*}
We call $P_t(Y,X)$ the active concept at time step $t$. Moreover, we distinguish between real and virtual concept drift. Virtual concept drift describes a change in $P(X)$, i.e. $P_t(X) \neq P_u(X)$. Hence, virtual concept drift is independent from the target distribution and does not change the decision boundary \cite{gama2014survey}. On the other hand, real concept drift, sometimes called concept shift, corresponds to a change in the conditional target distribution, i.e. $P_t(Y|X) \neq P_u(Y|X)$. Real concept drift shifts the decision boundary, which may influence subsequent predictions \cite{gama2014survey}. It is therefore crucial to detect changes of $P(Y|X)$ in time to avoid dramatic drops in predictive performance. In this paper, we investigate the effective and robust identification of real concept drift.

Unfortunately, concept drift does not follow a clear pattern in practice. Instead, we might observe large differences in the duration and magnitude of concept drift. To this end, we distinguish between different types of concept drift \cite{webb2016characterizing,gama2014survey,vzliobaite2010learning}: Sudden drift describes an abrupt change from one concept to another. Incremental drift is a steady transition of concepts over some time period. In a gradual drift, the concepts alternate temporarily, until a new concept ultimately replaces the old one. Sometimes we also observe mixtures of different concept drift types and recurring or cyclic concepts. For further information, we refer the fellow reader to \cite{webb2016characterizing}. In general, concept drift detection models should allow timely and accurate detection of all types of concept drift.

In a data stream, we can only access a fraction of the data at every time step $t$. To detect real concept drift, we thus need to approximate $P_t(Y|X)$, by using a predictive model $f_{\theta_t}$. Accordingly, we get $P_t(Y|X) \approx P(Y|X,\theta_t)$, with parameters $\theta_t =  (\theta_{tk})^K_{k=1}$. We optimize the model parameters, given the new observations in every time step. Consequently, $\theta_t$ represents our most current information about the active concept at time step $t$. A concept drift detection model should therefore adhere to changes of the model parameters through the following two properties:
\begin{property}
\label{prop:awareness}
    \textbf{Model-Aware Concept Drift Detection.} Let $\theta_t, \theta_u$ be the parameters of a predictive model $f_\theta$ at two time steps $t$ and $u$. Let further $\mathcal{D}$ be a statistical divergence measure (e.g., Kullback–Leibler, Jensen-Shannon, etc.). Concept drift detection is \emph{model-aware}, if for a detected drift between any two time steps $t$ and $u$, we observe $\mathcal{D}(\theta_t, \theta_u) > 0$.
\end{property}
Accordingly, we associate concept drift with updates of the predictive model $f_\theta$. Given that $f_\theta$ is robust, model-awareness reduces the sensitivity of a concept drift detection scheme to random input perturbations, which in turn reduces the risk of false alarms.

\begin{property}
\label{prop:explainability}
    \textbf{Explainable Concept Drift Detection.} Concept drift detection at time step $t$ is \emph{explainable} with respect to the predictive model $f_{\theta_t}$, if the concept drift can be associated with individual model parameters, i.e. each dimension of $\theta_t$.
\end{property}
If we associate concept drift with individual parameters, we can make more targeted model updates. Hence, we may avoid unnecessary and costly adaptations of the predictive model. Moreover, some parameter distributions even allow us to relate concept drift to specific input features. In this way, concept drift becomes much more transparent.

In this paper, we propose a novel framework for \emph{\underline{E}ffective and \underline{R}obust \underline{I}dentification of \underline{C}oncept \underline{S}hift (ERICS)}. \erics\ complies with the Properties \ref{prop:awareness} and \ref{prop:explainability}. We use the probabilistic framework introduced in \cite{haug2020leveraging} to model the distribution of the parameters $\theta$ at every time step. Specifically, we express real concept drift in terms of the marginal likelihood and the parameter distribution $P(\theta;\psi)$, which is itself parameterized by $\psi$. Unlike many existing models, \erics\ does not need to access the streaming data directly \cite{zhao2020handling}. Instead, we detect concept drift by investigating the differential entropy and Kullback-Leibler (KL) divergence of $P(\theta;\psi)$ at different time steps. In this context, we show that concept drift corresponds to changes in the distributional uncertainty of model parameters. In other words, real concept drift can be measured as a change in the average number of bits required to encode the parameters of the predictive model. By specifying an adequate parameter distribution, we can identify concept drift at the input level, which offers a significant advantage over existing approaches in terms of explainability. In fact, the proposed framework can be applied to almost any parameter distribution and online predictive model. For illustration, we apply \erics\ to a Probit model. In experiments on both synthetic and real-world data sets, we show that the proposed framework can detect different types of concept drift, while having a lower average delay than state-of-the-art methods. Indeed, \erics\ outperforms existing approaches with respect to the recall and precision of concept drift alerts.

In summary, we propose a generic and flexible framework that leverages the uncertainty patterns of model parameters for more effective concept drift detection in data streams. An open source version of \erics\ is available at https://github.com/haugjo/erics.

\section{ERICS: A Concept Drift Detection Framework}
Real concept drift corresponds to a change of the conditional target distribution $P(Y|X)$ \cite{webb2016characterizing}. However, data streams are potentially infinite and so the true distribution $P(Y|X)$ remains unknown. Hence, we may use a predictive model $f_\theta$ to approximate $P(Y|X)$. Since we update the model parameters $\theta$ for every new observation, $\theta_t$ represents our most current information about the active concept at time step $t$. Consequently, we may identify concept drift by investigating changes in $\theta$ over time.

To this end, we adopt the general framework of \cite{haug2020leveraging} and treat the parameters $\theta$ as a random variable, i.e. $\theta \sim P(\theta;\psi)$. Analogously, we optimize the distribution parameters $\psi$ at every time step with respect to the log-likelihood. This optimization problem can be expressed in terms of the marginal likelihood $P(Y|X,\psi)$ \cite{haug2020leveraging}. Hence, the marginal likelihood relates to the optimal parameter distribution under the active concept. Accordingly, we may associate concept drift between two time steps $t$ and $u$ with a difference of the marginal likelihood for the distribution parameters $\psi_t$ and $\psi_u$:
\begin{align}
\label{eq:drift}
    &P(Y|X;\psi_t) \neq P(Y|X;\psi_u)  \nonumber \\
    \Leftrightarrow |&P(Y|X;\psi_t) - P(Y|X;\psi_u)| > 0 \nonumber \\
    \Leftrightarrow \Big|&\int P(Y|X,\theta) \big[P(\theta;\psi_t) - P(\theta;\psi_u)\big] ~d\theta \Big| > 0.
\end{align}
From \eqref{eq:drift}, we may obtain a general scheme for concept drift detection. To this end, we rephrase \eqref{eq:drift} in terms of the differential entropy and KL-divergence, which are common measures from information theory. The entropy of a random variable corresponds to the average degree of uncertainty of the possible outcomes. Besides, entropy is often described as the average number of bits required to encode a sample of the distribution. On the other hand, the KL-divergence measures the difference between two probability distributions. It is frequently applied in Bayesian inference models, where it describes the information gained by updating from a prior to a posterior distribution. We can derive the following proportionality:
\begin{align}
\label{eq:entropy_drift}
    &\int P(Y|X,\theta) \big[P(\theta;\psi_t) - P(\theta;\psi_u)\big] ~d\theta \nonumber\\
    \propto &\int P(\theta;\psi_t) ~d\theta - \int P(\theta;\psi_u) ~d\theta \nonumber\\
    \propto &\int P(\theta;\psi_t) \log P(\theta;\psi_t) ~d\theta - \int P(\theta;\psi_u) \log P(\theta;\psi_t) ~d\theta \nonumber\\
    = &~H[P(\theta;\psi_u),P(\theta;\psi_t)] - h[P(\theta;\psi_t)] \nonumber\\
    = &~h[P(\theta;\psi_u)] - h[P(\theta;\psi_t)] + D_{KL}[P(\theta;\psi_u)\|P(\theta;\psi_t)],
\end{align}
where $h[P(\theta;\psi_t)]$ is the differential entropy of the parameter distribution at time step $t$. Note that we have rephrased the cross entropy $H[P(\theta;\psi_u),P(\theta;\psi_t)]$ by using the KL-divergence $D_{KL}$. We may now substitute \eqref{eq:entropy_drift} into \eqref{eq:drift} to derive a general scheme for concept drift detection:
\begin{equation}
\label{eq:info_drift}
    \big| \underbrace{h[P(\theta;\psi_u)] - h[P(\theta;\psi_t)]}_{\Delta \text{Uncertainty}} + \underbrace{D_{KL}[P(\theta;\psi_u)\|P(\theta;\psi_t)]}_{\Delta \text{Distribution}} \big| > 0
\end{equation}
Intuitively, real concept drift thus corresponds to a change in the uncertainty of the optimal parameters and a divergence of the parameter distribution. On the other hand, stable concepts are characterized by a static parameter distribution and uncertainty.

Note that \eqref{eq:info_drift} has another interpretation in the context of Bayesian inference. As mentioned before, the KL-divergence $D_{KL}[P(\theta;\psi_u)\|P(\theta;\psi_t)]$ can be interpreted as the information gained from inferring the posterior $P(\theta;\psi_u)$ from a prior $P(\theta;\psi_t)$. According to \eqref{eq:info_drift}, we thus find that every difference in parameter uncertainty (entropy) between time step $t$ and $u$, which can not be attributed to the inference of posterior parameters, may be traced back to a concept drift.

Finally, we show that the proposed concept drift detection scheme adheres to the Properties \ref{prop:awareness} and \ref{prop:explainability}.
\begin{IEEEproof}[Proof that ERICS is model-aware (Property \ref{prop:awareness})]
    By construction, we model the parameters $\theta$ through a distribution $P(\theta;\psi)$. According to \eqref{eq:info_drift}, we write
    \begin{equation*}
        \Big| \int P(\theta;\psi_t) \log P(\theta;\psi_t) ~d\theta - \int P(\theta;\psi_u) \log P(\theta;\psi_t) ~d\theta \Big|,
    \end{equation*}
    which is 0 iff $P(\theta;\psi_t) = P(\theta;\psi_u)$. Consequently, we find that Equation \eqref{eq:info_drift} evaluates to true, iff $P(\theta;\psi_t) \neq P(\theta;\psi_u)$. By definition, for any sensible statistical divergence measure $\mathcal{D}$, we know that $\mathcal{D}(P(\theta;\psi_t),P(\theta;\psi_u)) = 0 \Leftrightarrow P(\theta;\psi_t) = P(\theta;\psi_u)$. Equation \eqref{eq:info_drift} holds true, and thus $P(\theta;\psi_t) \neq P(\theta;\psi_u) \Leftrightarrow \mathcal{D}(P(\theta;\psi_t),P(\theta;\psi_u)) > 0$
\end{IEEEproof}

\begin{IEEEproof}[Proof that ERICS is explainable (Property \ref{prop:explainability})]
    By construction, any parametric distribution $P(\theta;\psi)$ used in Equation \eqref{eq:info_drift} can be evaluated for each parameter individually, i.e. we have $P(\theta_k;\psi_k)~\forall k$.
\end{IEEEproof}

\subsection{Continuous Concept Drift Detection}
Based on the general scheme \eqref{eq:info_drift}, we are able to identify concept drift between any two time steps $t$ and $u$. In practice, we are mainly interested in concept drifts between successive time steps $t-1$ and $t$. However, if we were to study \eqref{eq:info_drift} for two time steps only, our concept drift detection model might become too sensitive to random variations of the predictive model. To be more robust, we examine the moving average of \eqref{eq:info_drift} instead. Specifically, we compute the moving average at time step $t$ over $M$ time steps as
\begin{align}
\label{eq:moving_average}
    \text{MA}_t = \frac{1}{M} \sum^t_{i=t-M+1} \Big( \big| &h[P(\theta;\psi_i)] - h[P(\theta;\psi_{i-1})] + \nonumber\\ 
    &D_{KL}[P(\theta;\psi_i)\|P(\theta;\psi_{i-1})] \big| \Big).
\end{align}
As before, the moving average contains our latest information on the model parameters and the active concept. We can adjust the sensitivity of our framework by selecting $M$ appropriately. In general, the larger we select $M$, the more robust the framework becomes. However, a large $M$ might also hide concept drifts of small magnitude or short duration.

So far we have treated all changes of the parameter distribution as an indication of concept drift. Indeed, this is in line with the general definition of concept drift \cite{webb2016characterizing}. Still, we argue that only certain changes in the parameter distribution have practical relevance. For example, suppose that we use stochastic gradient descent (SGD) to optimize the model parameters at every time step. If we start from an arbitrary initialization, the distribution of optimal parameters usually changes significantly in early training iterations. However, given that the concept $P(Y|X)$ is stationary, SGD will almost surely converge to a local optimum. Consequently, we will ultimately minimize the entropy and KL-divergence of $P(\theta;\psi)$ in successive time steps. In other words, \eqref{eq:moving_average} will tend to decrease as long as we optimize the parameters $\psi$ with respect to the active concept. However, if the decision boundary changes due to a real concept drift, SGD-updates will aim for a different optimum. This change of the objective will temporarily lead to more uncertainty in the model and thus increase the entropy of the parameter distribution. 

\begin{figure}[t]
\centering
     \subfloat[$\beta = 0.01$]{\includegraphics[width=0.9\columnwidth]{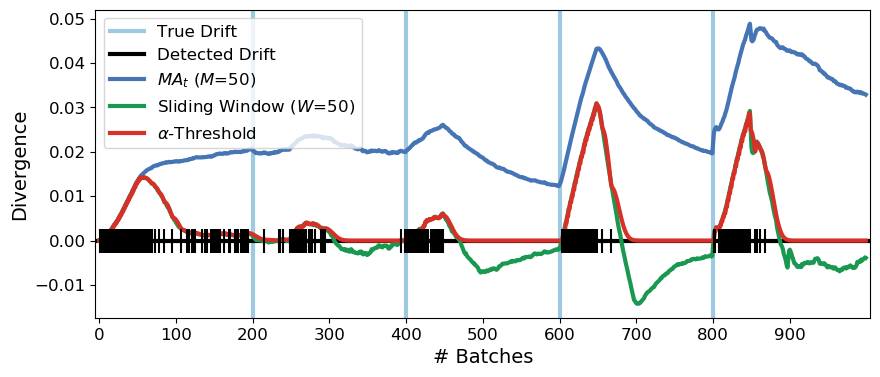}}
     \hfill
     \subfloat[$\beta = 0.001$]{\includegraphics[width=0.9\columnwidth]{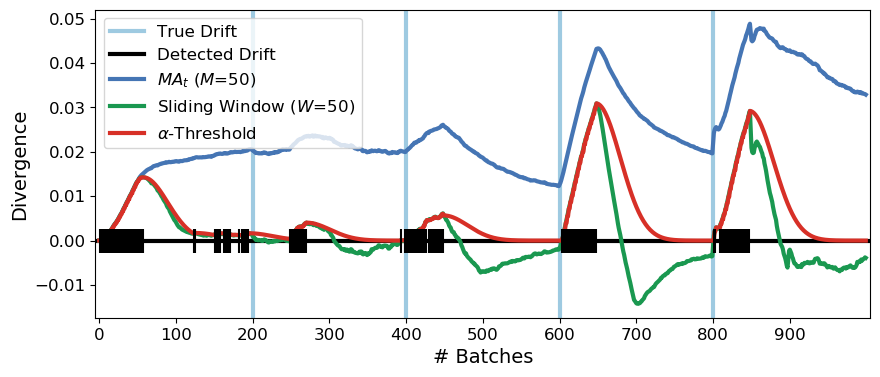}}
     \hfill
     \subfloat[$\beta = 0.0001$]{\includegraphics[width=0.9\columnwidth]{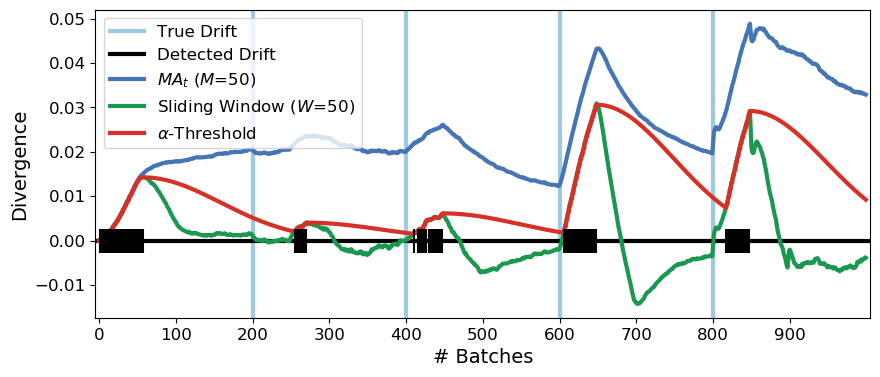}}
    \caption{\emph{Updating the $\alpha$-Threshold.} The proposed framework uses a dynamic threshold $\alpha$ (red line) to detect concept drift. According to \eqref{eq:moving_average}, we track a moving average of the divergence of the parameter distribution (dark blue line). If the total divergence in a sliding window (green line) exceeds the threshold, \erics\ detects a concept drift (black vertical lines). By adjusting the hyperparameter $\beta$, we can control the iterative updates of $\alpha$ and thus regulate the sensitivity of \erics\ after a drift is detected. Generally, the larger we choose $\beta$, the more sensitive \erics\ becomes to changes of the parameter distribution. Here, we depict different $\beta$ for the KDD data set \cite{Dua2019}. We artificially generated four sudden concept drifts (blue vertical lines). In this example, small update steps (i.e. small $\beta$) are preferable to give the predictive model enough time to adapt to the new concept. Note that the early alerts correspond to the initial training phase of the predictive model. Hence, we would ignore them in practice.}
    \label{fig:alpha_update}
\end{figure}

We exploit this temporal pattern for concept drift detection. To this end, we measure the total change of \eqref{eq:moving_average} in a sliding window of size $W$:
\begin{equation}
\label{eq:sliding_window}
    \sum^t_{j=t-W+1} \big( \text{MA}_j - \text{MA}_{j-1} \big) > \alpha_t \Leftrightarrow \text{Drift at $t$},
\end{equation}
where $\alpha_t \geq 0$ is an adaptive threshold. As before, we may control the robustness of the concept drift detection with the sliding window size $W$. Whenever we detect concept drift, i.e. \eqref{eq:sliding_window} evaluates to true, we redefine $\alpha_t$ as
\begin{equation}
\label{eq:alpha}
    \alpha_t = \sum^t_{j=t-W+1} \big( \text{MA}_j - \text{MA}_{j-1} \big).
\end{equation}
In this way, we temporarily tolerate all changes to the predictive model up to a magnitude of \eqref{eq:alpha}. We consider these changes to be the after-effects of the concept drift. We then update $\alpha_t$ in an iterative fashion. Let $\beta$ be a user-defined hyperparameter in the interval $[0,1]$. Each update depends on the current $\alpha$-value, the $\beta$-hyperparameter and the time elapsed since the last concept drift alert, which we denote by $\Delta_{Drift}$:
\begin{equation}
\label{eq:alpha_update}
    \alpha_t = \alpha_{t-1} - (\alpha_{t-1} * \beta * \Delta_{Drift})
\end{equation}
Note that $\alpha_t$ will asymptotically approach 0 over time, if there is no concept drift. In this way, we gradually reduce the tolerance of our framework after a drift is detected.

The choice of a suitable $\beta$ usually depends on the application at hand. By way of illustration, we applied \erics\ with different $\beta$ to the KDD data set \cite{Dua2019}. We used \cite{sethi2017reliable}'s method to induce sudden concept drift after every 20\% of observations. For more information, see Section \ref{sec:experiments}. Figure \ref{fig:alpha_update} illustrates the components of \erics\ for three different $\beta$-values. Notably, the larger we chose $\beta$, the more drifts we detected. Since we were dealing with a sudden concept drift in this particular example, we could be less sensitive and apply smaller update steps. For $\beta=0.0001$, we achieved good first results in all our experiments. Therefore, this value can generally be used as a starting point for further optimization.

To conclude our general framework, we provide a pseudo code implementation in Figure \ref{alg:erics}.

\subsection{Limitations and Advantages}
The proposed framework does not access streaming observations directly, but uses the parameters of a predictive model instead. Accordingly, our approach is much more memory efficient than many related works. Yet, if the parameter distribution does not change in a drift period, concept drift may go unnoticed. In general, however, \erics\ can detect all concept drifts that affect the predictive outcome.

One should also be aware that some predictive models are prone to adversarial attacks. Accordingly, \erics\ can only be as robust as its underlying predictive model. This sensitivity to the predictive model is shared by most existing works. With \erics, the possibility of misuse is drastically reduced, as we closely monitor the distribution of the model parameters at all times.

\begin{figure}[t]
    \begin{algorithmic}
        \REQUIRE $[\psi_t,..,\psi_{t-M}]$; $[\text{MA}_{t-1},..,\text{MA}_{t-W}]$; $\alpha_{t-1}$; $\Delta_{Drift}$
        \STATE $\alpha_t \leftarrow Eq.~\eqref{eq:alpha_update}$
        \STATE $\Delta_{Drift} \leftarrow \Delta_{Drift} + 1$
        \STATE $\text{MA}_t \leftarrow Eq.~\eqref{eq:moving_average}$
        \STATE $sumWindow \leftarrow \sum^t_{j=t-W+1} \big( \text{MA}_j - \text{MA}_{j-1} \big)$
        \STATE
        \IF{$sumWindow > \alpha_t$}
        \STATE $\alpha_t \leftarrow sumWindow$
        \STATE $\Delta_{Drift} \leftarrow 1$
        \ENDIF
        \STATE
        \RETURN $\alpha_t$; $\text{MA}_t$; $\Delta_{Drift}$
    \end{algorithmic}
\caption{\emph{Pseudo Code.} Concept drift detection with \erics\ at time step $t$.}
\label{alg:erics}
\end{figure}

\section{Illustrating ERICS}
\erics\ is model-agnostic. This means that the framework can be applied to different predictive models $f_\theta$ and parameter distributions $P(\theta;\psi)$. In this way, we enable maximum flexibility with regard to possible streaming applications. By way of illustration, we adopt a Probit model with independent normally distributed parameters. This setup has achieved state-of-the-art results in online feature selection \cite{haug2020leveraging}. Besides, it offers dramatic computational advantages due to its low complexity. In line with \cite{haug2020leveraging}, we optimize $\psi$ at every time step with respect to the log-likelihood for the Probit model.

The assumption of independent model parameters may appear restrictive, but in practice it often leads to good results, e.g. in the case of local feature attributions \cite{lundberg2017unified,kasneci2016licon} or feature selection \cite{haug2020leveraging,borisov2019cancelout}. In fact, the independence assumption allows us to identify the parameters affected by concept drift and thus to comply with Property \ref{prop:explainability}. Since the Probit model comprises one parameter per input feature, we can readily associate concept drift with individual input variables.

Accordingly, let $P(\theta;\psi_t) = \mathcal{N}(\psi_t=(\mu_t, \Sigma_t))$, where $\mu_t = (\mu_{tk})^K_{k=1}$ is a vector of mean values and $\Sigma_t$ is the diagonal covariance matrix, where the diagonal entries correspond to the vector $\sigma^2_t = (\sigma^2_{tk})^K_{k=1}$. The differential entropy of $P(\theta;\psi_t)$ is
\begin{equation*}
    h\big[P(\theta;\psi_t)\big] = \frac{1}{2} \Big( K + K \ln (2\pi) + \ln \prod^K_{k=1} \sigma^2_{tk} \Big).
\end{equation*}
The KL-divergence between $P(\theta;\psi_t)$ and $P(\theta;\psi_{t-1})$ is
\begin{align*}
    &D_{KL}[P(\theta;\psi_t)\|P(\theta;\psi_{t-1})] \\
    = &\frac{1}{2} \left(\sum^K_{k=1} \frac{\sigma^2_{tk} + (\mu_{t-1,k} - \mu_{tk})^2}{\sigma^2_{t-1,k}} - K + \ln \frac{\prod^K_{k=1} \sigma^2_{t-1,k}}{\prod^K_{k=1} \sigma^2_{tk}} \right).
\end{align*}
According to \eqref{eq:moving_average}, we then write the moving average as
\begin{equation}
\label{eq:drift_probit}
    \text{MA}_t = \frac{1}{2M} \sum^t_{i=t-M+1} \Bigg|\sum^K_{k=1} \frac{\sigma^2_{ik} + (\mu_{i-1,k} - \mu_{ik})^2}{\sigma^2_{i-1,k}} - K \Bigg|. 
\end{equation}
Note that \eqref{eq:drift_probit} scales linearly with the number of parameters $K$, i.e. it has $\mathcal{O}(K)$ time complexity.

In order to identify concept drift at individual parameters (which is equivalent to examining individual features, since we use a Probit model), we can investigate the moving average of a specific parameter $\theta_k$:
\begin{equation}
\label{eq:partial_drift}
    \text{MA}_{tk} = \frac{1}{2M} \sum^t_{i=t-M+1} \Bigg| \frac{\sigma^2_{ik} + (\mu_{i-1,k} - \mu_{ik})^2}{\sigma^2_{i-1,k}} - 1 \Bigg|
\end{equation}
In this case, we maintain a different threshold $\alpha_k$ per parameter. Note that \eqref{eq:partial_drift} has a constant time complexity.

\section{Related Work}\label{sec:related_work}
In this section, we briefly introduce some of the most prominent and recent contributions to concept drift detection. 

DDM monitors changes in the classification error of a predictive model \cite{gama2004learning}. Whenever the observed error changes significantly, DDM issues a warning or an alert. We find various modifications of this general scheme, including \cite{baena2006early} and \cite{barros2017rddm}. Another well-known method for concept drift adaptation is ADWIN \cite{bifet2007learning}. Here, the authors maintain a sliding window, whose size changes dynamically according to the current rate of distributional change. \cite{de2018concept} also employ a sliding window approach and provide a feasible implementation of Fisher's Exact test, which they use for concept drift detection. Similar to our framework, \cite{du2014detecting} use a sliding window and the entropy to detect concept drift. However, they examine entropy with regard to the predictive result and disregard the model parameters. FHDDM applies a sliding window to classification results and tracks significant differences between the current probability of correct predictions and the previously observed maximal probability \cite{pesaranghader2016fast}. To this end, FHDDM employs a threshold that is based on the Hoeffding bound. In a later approach, the same authors instead use McDiarmid's inequality to detect concept drift \cite{pesaranghader2018mcdiarmid}. EWMA is a method that monitors an increase in the probability that observations are misclassified \cite{ross2012exponentially}. The authors use an exponentially weighted moving average, which places greater weight on the most recent instances in order to detect changes. \cite{harel2014concept} also focus on the predictive outcome. Specifically, they investigate the distribution of the loss function via resampling. Likewise, the LFR method uses certain test statistics to detect concept drift by identifying changes through statistical hypothesis testing \cite{wang2015concept}. Finally, \cite{sobhani2011new} compare the labels of close data points in successive batches to detect concept drift.

In addition, we find various approaches that examine ensembles of online learners to deal with concept drift. For example, \cite{bach2008paired} compare two models; one that is trained with all streaming observations and another that is trained only with the latest observations. Likewise, \cite{tan2019online} analyze the density of the posterior distributions of an incremental and a static estimator. 

More information about the progress in concept drift detection can be found in \cite{vzliobaite2010learning,gama2014survey,webb2016characterizing,gonccalves2014comparative}.

Conceptually, our work differs substantially from the remaining literature.
Instead of directly examining the streaming observations or the predictive outcome, \erics\ monitors changes in the parameters of a predictive model.

\section{Experiments}
\label{sec:experiments}
We evaluated \erics\ in multiple experiments. All experiments were conducted on an i5-8250U CPU with 8 Gb of RAM, running 64-bit Windows 10 and Python 3.7.3. We compared our framework to the popular concept drift detection methods ADWIN \cite{bifet2007learning}, DDM \cite{gama2004learning}, EWMA \cite{ross2012exponentially}, FHDDM \cite{pesaranghader2016fast}, MDDM \cite{pesaranghader2018mcdiarmid} and RDDM \cite{barros2017rddm}. We used the predefined implementations of these models as provided by the Tornado framework \cite{pesaranghader2018reservoir}. Besides, we applied the default set of parameters throughout all experiments. Note that all related models require classifications of a predictive model. To this end, we trained a Very Fast Decision Tree (VFDT) \cite{hulten2001mining} in an interleaved test-then-train evaluation. The VFDT is a state-of-the-art online learner, which uses the Hoeffding bound to incrementally construct a decision tree for streaming data. We used the VFDT implementation of scikit-multiflow \cite{montiel2018scikit} in our experiments. Note that we consider a simple binary classification scenario in all our experiments, since it should be handled well by all models.

We optimized the hyperparameters of \erics\ in a grid search. The search space was either chosen empirically or according to \cite{haug2020leveraging}. Table \ref{tab:hyperparameters} lists all hyperparameters per data set. The hyperparameters ``Epochs'', ``LR (learning rate) $\mu$'' and ``LR $\sigma$'' control the training of the Probit model, which we adopted from \cite{haug2020leveraging}.

\begin{table}[t]
\caption{Synthetic and Real World Data Sets}
    \label{tab:datasets}
    \centering
        \begin{tabular}{llll}
        \toprule
        Name & \#Samples & \#Features & Data Types\\ 
        \cmidrule(lr){1-1} \cmidrule(lr){2-4}
        SEA (synth.) & 100,000 & 3 & cont.\\
        Agrawal (synth.) & 100,000 & 9 & cont.\\
        Hyperplane (synth.) & 100,000 & 20 & cont.\\
        Mixed (synth.) & 100,000 & 9 & cont.\\
        Spambase & 4,599 & 57 & cont.\\
        Adult & 48,840 & 54 & cont., cat.\\
        HAR (binary) & 7,450 & 562 & cont.\\
        KDD (sample) & 100,000 & 41 & cont., cat.\\
        Dota & 102,944 & 116 & cat.\\
        MNIST (binary) & 10,398 & 784 & cont.\\
        \bottomrule
        \end{tabular}
\end{table}

\begin{table}[t]
\caption{Hyperparameters of ERICS per Data Set}
    \label{tab:hyperparameters}
    \centering
        \begin{tabular}{llllllll}
        \toprule
        Data Set & $M$ & $W$ & $\beta$ & Epochs & LR $\mu$ & LR $\sigma$\\ 
        \cmidrule(lr){1-1} \cmidrule(lr){2-4}\cmidrule(lr){5-7}
        SEA & 75 & 50 & 0.0001 & 10 & 0.01 & 0.01\\
        Agrawal & 100 & 50 & 0.001 & 10 & 0.01 & 0.01\\
        Hyperplane & 100 & 50 & 0.0001 & 10 & 0.01 & 0.01\\
        Mixed & 100 & 50 & 0.0001 & 10 & 0.1 & 0.01\\
        Spambase & 35 & 25 & 0.001 & 10 & 0.1 & 0.01\\
        Adult & 50 & 50 & 0.001 & 10 & 0.1 & 0.01\\
        HAR & 25 & 50 & 0.001 & 10 & 0.1 & 0.01\\
        KDD & 50 & 50 & 0.0001 & 10 & 0.01 & 0.01\\
        Dota & 75 & 50 & 0.0001 & 10 & 0.01 & 0.01\\
        MNIST & 25 & 20 & 0.001 & 50 & 0.1 & 0.01\\
        \bottomrule
        \end{tabular}
\end{table}

\subsection{Data Sets}
In order to evaluate the timeliness and precision of a concept drift detection model, we require ground truth. Consequently, we generated multiple synthetic data sets using the scikit-multiflow package \cite{montiel2018scikit}. Detailed information about each generator can be obtained from the corresponding documentation. We exhibit the properties of all data sets in Table \ref{tab:datasets}. Note that we simulated multiple types of concept drift. Specifically, we produced sudden concept drifts with the SEA generator. To this end, we specified a drift duration (\emph{width} parameter) of 1. We alternated between the classification functions 0-3 to produce the different concepts. With the Agrawal generator, we simulated gradual drift of different duration. Again, we alternated between the classification functions 0-3 to shift the data distribution. With the rotating Hyperplane generator, we simulated an incremental drift over the full length of the data set. We generated 20 features with the Hyperplane generator, out of which 10 features were subject to concept drift by a magnitude of 0.5. Finally, we produced a Mixed drift using the Agrawal generator. The Mixed data contains both sudden and gradual drift, which we obtained by alternating the classification functions 0-4. All synthetic data sets contain 10\% noisy data. We obtained 100,000 observations from each data stream generator.

In addition, we evaluated the proposed framework on real world data. However, since real world data usually does not provide any ground truth information, we had to artificially induce concept drift. For this reason, we applied the methodology of \cite{sethi2017reliable} to induce sudden concept drift in five well-known data sets from the online learning literature. First, we randomly shuffled the data to remove any natural (unknown) concept drifts. Next, we ranked all features according to their information gain. We then selected the top 50\% of the ranked features and randomly permuted their values. In this way, we generated sudden drifts after every 20\% of the observations. Specifically, we introduced concept drift to the real-world data sets Spambase, Adult, Human Activity Recognition (HAR), KDD 1999 and Dota2, which we took from the UCI Machine Learning repository \cite{Dua2019}. Note that we drew a random sample of 100,000 observations from the KDD 1999 data to allow for feasible computations.

Besides, we used the MNIST data set to evaluate partial concept drift detection at the input level. We selected all observations that are either labelled \emph{3} or \emph{8}, since these numbers are difficult to distinguish. In the first half of the observations, we treated \emph{3} as the true class. In the second half of the observations, we switched the true class to \emph{8}. In this way, we simulated a sudden concept drift of all input features.

For all real world data sets, we normalized the continuous features to the range $[0,1]$ and one-hot-encoded the categorical features. In the Adult data set, we imputed all NaN-values by a new category \emph{unknown}. Moreover, we altered the labels of the HAR data set to simulate binary classification between the class \emph{moving} (original labels \emph{walking}, \emph{walking\_downstairs} and \emph{walking\_upstairs}) and \emph{non-moving} (original labels \emph{sitting}, \emph{laying} and \emph{standing}). We trained the online predictive models (Probit and VFDT) in batches of the following size: For Spambase and HAR we chose a batch size of 10. Adult was processed in batches of size 50. For all remaining data sets, we trained on batches of 100 observations.

\begin{figure*}[ht]
\centering
     \subfloat[SEA]{\includegraphics[width=0.3\textwidth]{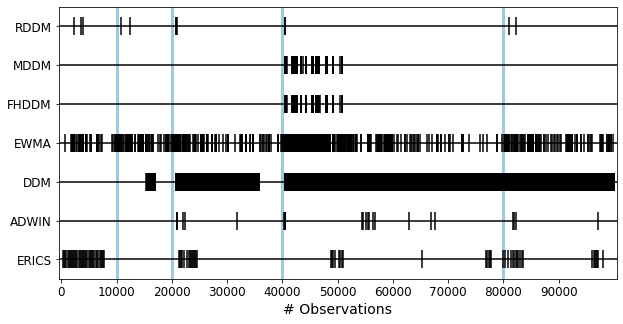}}
     \hfill
     \subfloat[Agrawal]{\includegraphics[width=0.3\textwidth]{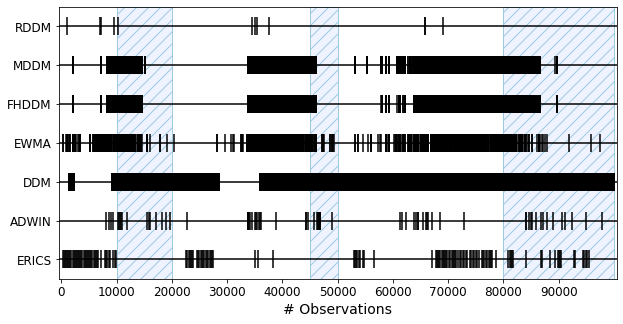}}
     \hfill
     \subfloat[Hyperplane]{\includegraphics[width=0.3\textwidth]{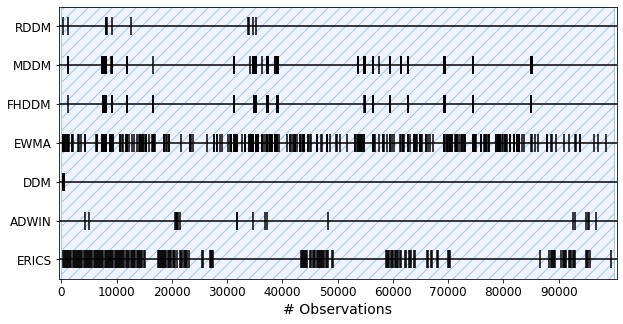}}
     \hfill
     \subfloat[Mixed]{\includegraphics[width=0.3\textwidth]{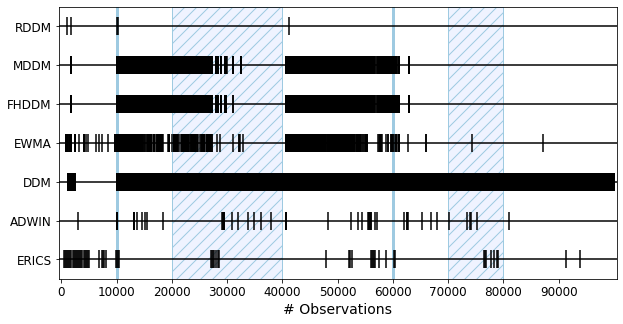}}
     \hfill
     \subfloat[Spambase]{\includegraphics[width=0.3\textwidth]{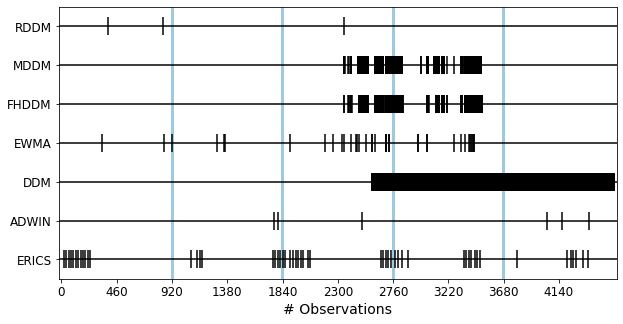}}
     \hfill
     \subfloat[Adult]{\includegraphics[width=0.3\textwidth]{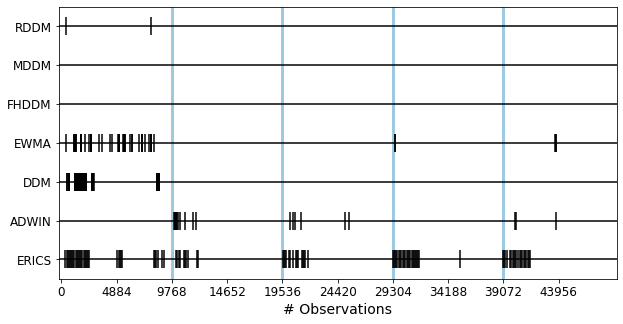}}
     \hfill
     \subfloat[HAR]{\includegraphics[width=0.3\textwidth]{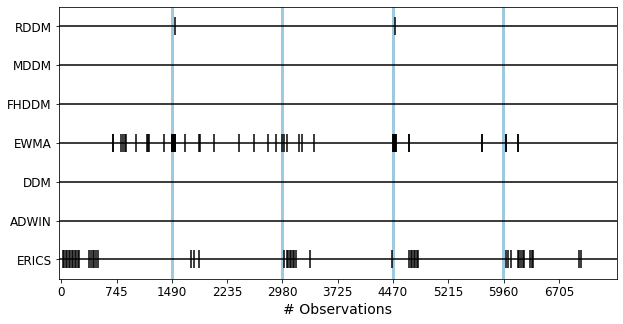}}
     \hfill
     \subfloat[KDD]{\includegraphics[width=0.3\textwidth]{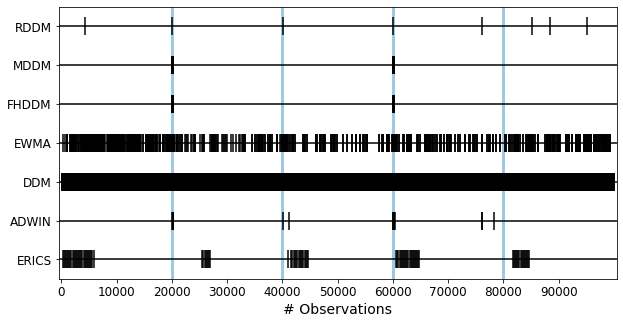}}
     \hfill
     \subfloat[Dota]{\includegraphics[width=0.3\textwidth]{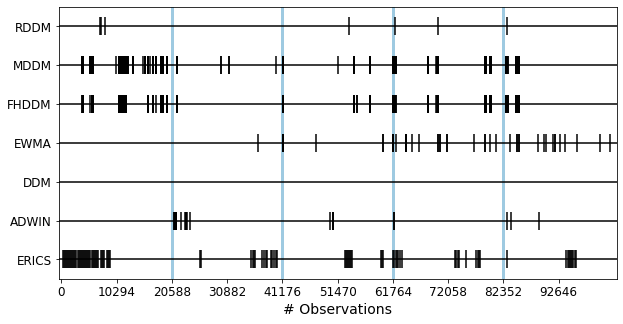}}
    \caption{\emph{Drifts Alerts.} For all data sets, we illustrate the drift alerts obtained from each concept drift detection model. The blue vertical lines and shaded areas correspond to known concept drifts. Each black marker stands for one drift alert. Early drift alerts can be attributed to the initial training of the predictive model and were therefore ignored. Notably, \erics\ (ours) seems to detect most concept drifts, while triggering considerably fewer false alarms than most related models. We find support for this intuition in the remaining figures.}
    \label{fig:drifts}
\end{figure*}

\begin{figure}[ht]
\includegraphics[width=0.9\columnwidth]{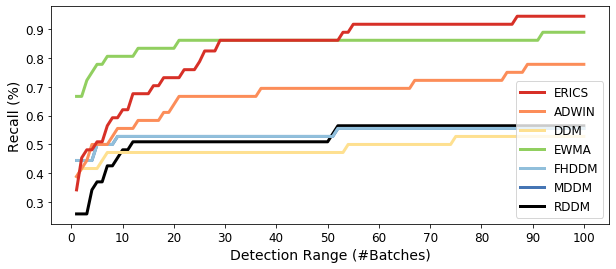}
\caption{\emph{Recall.} We show the average recall over all data sets for different detection ranges. \erics\ (ours) ultimately detects more than 90\% of the known concept drifts. The apparent disadvantage of \erics\ in early batches can be attributed to the slower update speed of the Probit model as compared to the VFDT \cite{hulten2001mining}, which was used by the remaining concept drift detection methods. Besides, the recall scores should be considered with care, since some methods tend to detect drift at almost every time step and are thus not reliable.}
\label{fig:recall}
\end{figure}

\begin{figure}[ht]
\includegraphics[width=0.9\columnwidth]{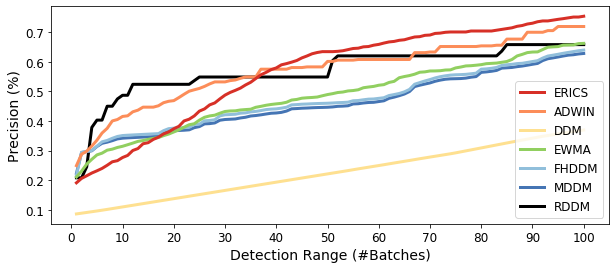}
\caption{\emph{Precision.} We show the average precision over all data sets for different detection ranges. \erics\ (ours) tends to identify concept drift later than some related models. Therefore, we count fewer true positives in small detection ranges, which leads to lower precision. However, for larger detection ranges, our framework is more precise than any other model in the evaluation.}
\label{fig:precision}
\end{figure}

\begin{table}[t]
\caption{Average Delay in Number of Batches}
    \label{tab:delay}
    \centering
    \begin{adjustbox}{max width=\columnwidth}
        \begin{tabular}{llllllllll}
        \toprule
        & \multicolumn{7}{c}{Drift Detection Models}\\
        \cmidrule(lr){1-1}\cmidrule(lr){2-8}
        Datasets & ERICS & ADWIN & DDM & EWMA & FHDDM & MDDM & RDDM\\
        \cmidrule(lr){1-1}\cmidrule(lr){2-8}
        SEA & 52.75 & 34.55 & 16.45 & \textbf{0.26} & 178.61 & 178.58 & 7.42\\
        Agrawal & 71.33 & 16.80 & \textbf{0.00} & 0.31 & 0.023 & 0.20 & 137.22\\
        Hyperplane & \textbf{2.00} & 42.27 & 2.23 & 2.36 & 11.24 & 11.22 & 2.42\\
        Mixed & 34.00 & 27.95 & \textbf{0.34} & 11.55 & 75.26 & 75.25 & 227.98\\
        Spambase & \textbf{7.35} & 79.0 & 60.23 & 29.96 & 71.63 & 71.58 & 117.45\\
        Adult & \textbf{2.61} & 63.15 & 488.39 & 172.57 & 488.39 & 488.39 & 488.39\\
        HAR & 13.05 & 372.45 & 372.45 & \textbf{1.55} & 372.45 & 372.45 & 77.10\\
        KDD & 22.01 & 50.35 & \textbf{0.00} & 0.55 & 100.25 & 100.21 & 13.21\\
        Dota & 44.04 & 25.32& 514.72 & 43.40 & 3.56 & \textbf{3.54} & 116.46\\
        \cmidrule(lr){1-8}
        Mean & \textbf{27.68} & 79.09 & 161.65 & 29.17 & 144.60 & 144.60 & 131.96\\
        Rank & \textbf{1} & 3 & 7 & 2 & 5 & 5 & 4\\
        \bottomrule
        \end{tabular}
    \end{adjustbox}
\end{table}

\subsection{Delay, Recall and Precision}
In our first experiment, we applied the concept drift detection models to all synthetic and real-world data sets. Figure \ref{fig:drifts} exhibits the drift alerts of every model. The blue vertical lines and shaded areas indicate periods of concept drift. Each black vertical line corresponds to one drift alert. Most models identify concept drift in early iterations. This is due to the initial training phase of the predictive model and therefore has no practical relevance. For the upcoming evaluations, we have therefore ignored all drift alerts in the first 80 batches.

By Figure \ref{fig:drifts}, the proposed framework \erics\ performs well in all data sets. Given the low complexity of the underlying Probit model, some concept drifts do not infer a change of the parameter distribution immediately. This can be seen in small delays, such as for the Agrawal data, for example. Still, \erics\ achieves the smallest average delay of all concept drift detection models, which is shown in Table \ref{tab:delay}.

\begin{figure}[t]
\includegraphics[width=0.9\columnwidth]{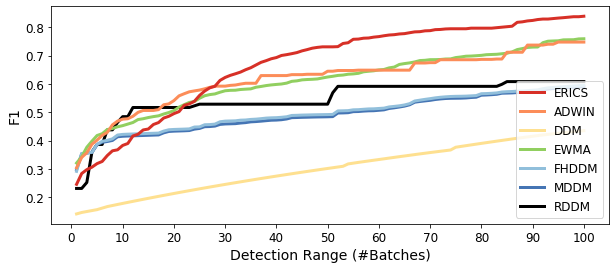}
\caption{\emph{F1:} We illustrate the F1 measure, which is the harmonic mean of the precision and recall shown in earlier plots. Here, the advantage of \erics\ (ours) is most apparent, since it significantly outperforms all related methods for a detection range greater than 30 batches.}
\label{fig:f1}
\end{figure}

Strikingly, \erics\ generally seems to produce fewer false alarms than related models. We find support for this intuition by examining the average recall (Figure \ref{fig:recall}) and precision (Figure \ref{fig:precision}) over all data sets. Similar to \cite{yu2017concept}, we evaluated the detected drifts for different detection ranges. The detection range corresponds to the number of batches after a known drift, during which we consider an alert as a true positive. Whenever there is no drift alert in the detection range, we count this as a false negative. Besides, all drift alerts outside of the detection range are false positives. We used these scores to compute the recall and precision values. Again, we find that \erics\ tends to struggle in the early stages, right after a drift happens. As mentioned before, we attribute this to the slowly updating Probit model that we used for illustration. The VFDT, which is used by all related models, is much more complex and can thus adapt to changes faster. Additionally, we must treat some recall scores with care. For example, in four data sets, the DDM model detects drift in almost every time step. Hence, it achieves perfect recall, although the drift alerts are not reliable at all. Still, \erics\ ultimately outperforms all related models in terms of both recall and precision. The superiority of our framework is even more apparent, if we look at the harmonic mean of precision and recall, which is the F1 score that we show in Figure \ref{fig:f1}.

\subsection{Detecting Drift at the Input Level}
\begin{figure}[t]
\includegraphics[width=\columnwidth]{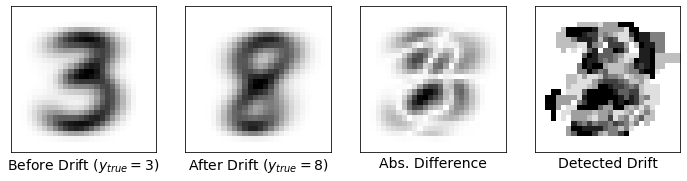}
\caption{\emph{Partial Drift Detection.} By choosing an appropriate base model and parameter distribution, \erics\ can attribute concept drift to individual input features. We selected all observations of MNIST with the label \emph{3} or \emph{8} and induced concept drift by changing the true class after half of the observations. In the left subplots, we exhibit the mean of the true class before and after the concept drift. The third subplot depicts the absolute difference of these mean values. In the right subplot, we show the alerts of \erics\ in the first 15 batches after the concept drift. The color intensity corresponds to the frequency of drift alerts per input feature. Strikingly, the drift alerts seem to map the absolute difference between both concepts. This suggests, that \erics\ does indeed identify concept drift for those input features that are most affected by a distributional change.}
\label{fig:mnist}
\end{figure}

As mentioned before, by using a Probit model and treating parameters as independently Gaussian distributed, we are able to associate concept drift with specific input features. By means of illustration, we apply \erics\ to a sample of the MNIST data set, which we induced with concept drift. In Figure \ref{fig:mnist}, we exhibit the mean of all observations corresponding to the true class before and after the concept drift (left subplots). We also show the absolute difference between those mean values. In the outer most subplot on the right, we illustrate the drift alerts per input feature in the first 15 batches after the concept drift. The color intensity corresponds to the number of drift alerts (where many alerts correspond to darker patterns). Strikingly, the frequency of drift alerts closely maps the absolute difference between the two concepts. This shows that \erics\ is generally able to identify the input features that are most affected by concept drift. We expect this pattern to become even clearer, when using more complex base models.

\section{Conclusion}
In this work, we proposed a novel and generic framework for the detection of concept drift in streaming applications. Our framework monitors changes in the parameters of a predictive model to effectively identify distributional changes of the input. We exploit common measures from information theory, by showing that real concept drift corresponds to changes of the uncertainty regarding the optimal parameters. Given an appropriate parameter distribution, the proposed framework can also attribute drift to specific input features. In experiments, we highlighted the advantages of our approach over multiple existing methods, using both synthetic and real-world data. Strikingly, \erics\ detects concept drift with less delay on average, while outperforming existing models in terms of both recall and precision.

\bibliographystyle{IEEEtran}
\bibliography{bibliography}

\end{document}